\documentclass[11pt,a4paper]{article}
\usepackage[hyperref]{acl2020}
\usepackage{times}
\usepackage{latexsym}

\usepackage[protrusion=true,expansion=true]{microtype}
\usepackage{subcaption}
\usepackage [english]{babel}
\usepackage [autostyle, english = american]{csquotes}
\MakeOuterQuote{"}
\usepackage{array, caption, floatrow, makecell}
\usepackage{booktabs, graphicx}

\newcolumntype{P}[1]{>{\centering\arraybackslash}p{#1}}
\usepackage{multirow}
\usepackage{color, colortbl}
\definecolor{LightCyan}{rgb}{0.88,1,1}
\definecolor{teagreen}{rgb}{0.82, 0.94, 0.75}
\definecolor{lightblue}{rgb}{0.68, 0.85, 0.9}
\definecolor{darkseagreen}{rgb}{0.56, 0.93, 0.56}
\definecolor{mygray}{gray}{0.45}
\definecolor{lightgray}{rgb}{0.90, 0.90, 0.90}

\newcommand\blfootnote[1]{%
	\begingroup
	\renewcommand\thefootnote{}\footnote{#1}%
	\addtocounter{footnote}{-1}%
	\endgroup
}

\aclfinalcopy 

\title{Schema-Guided Natural Language Generation}

\author{
	\bf Yuheng Du$^{*}$, Shereen Oraby$^{*}$, Vittorio Perera$^{*}$, Minmin Shen, Anjali Narayan-Chen, \\
	\bf Tagyoung Chung, Anu Venkatesh, Dilek Hakkani-Tur \\
	Amazon Alexa AI \\
	\texttt{\{yuhendu,orabys,pererv,shenm,naraanja,}\\
	\texttt{tagyoung,anuvenk,hakkanit\}@amazon.com}
}
\date{}

\begin{document}
\maketitle
\begin{abstract}
Neural network based approaches to data-to-text natural language generation (NLG) have gained popularity in recent years, with the goal of generating a natural language prompt that accurately realizes an input meaning representation. To facilitate the training of neural network models, researchers created large datasets of paired utterances and their meaning representations. However, the creation of such datasets is an arduous task and they mostly consist of simple meaning representations composed of slot and value tokens to be realized. These representations do not include any contextual information that an NLG system can use when trying to generalize, such as domain information and descriptions of slots and values. In this paper, we present the novel task of Schema-Guided Natural Language Generation (SG-NLG). Here, the goal is still to generate a natural language prompt, but in SG-NLG, the input MRs are paired with rich schemata providing contextual information. To generate a dataset for SG-NLG we re-purpose an existing dataset for another task: dialog state tracking, which includes a large and rich schema spanning multiple different attributes, including information about the domain, user intent, and slot descriptions. We train different state-of-the-art models for neural natural language generation on this dataset and show that in many cases, including rich schema information allows our models to produce higher quality outputs both in terms of semantics and diversity. We also conduct experiments comparing model performance on seen versus unseen domains, and present a human evaluation demonstrating high ratings for overall output quality.
\end{abstract}

\section{Introduction}
\label{sec:intro}
\blfootnote{$^{*}$Authors contributed equally and are listed alphabetically.}Much of the recent work on Neural Natural Language Generation (NNLG) focuses on generating a natural language string given some input content, primarily in the form of a structured Meaning Representation (MR)  \cite{moryossef-etal-2019-step,wiseman2017challenges,Gong2019TabletoTextGW,findingse2e,Liu2017TabletotextGB,webnlg_challenge,Wen2016MultidomainNN,DBLP:journals/corr/DusekJ16a,Dusek2015,Wen15}. Popular datasets used for MR-to-text generation are confined to limited domains, e.g., restaurants or product information, and usually consist of simple tuples of slots and values describing the content to be realized, failing to offer any information about domains or slots that might be useful to generation models \cite{e2edataset,webnlg,Wen15}. Table~\ref{table:traditional-MR} shows examples of MRs from popular datasets. 

\begin{table}[h!]
	\begin{footnotesize}
		\begin{tabular}
			{@{} p{1.2cm}|p{3cm}p{2.5cm} @{}}
			\toprule
			{\bf  Dataset } & {\bf  MR } & {\bf Reference }  \\ \midrule
			{\bf E2E \cite{e2edataset}} & { INFORM name[The Punter], food[Indian], priceRange[cheap]} & The Punter offers cheap Indian food. \\ \hline
			{\bf Laptop \cite{Wen2016MultidomainNN}} & { INFORM name[satellite eurus65], type[laptop], memory[4gb], driverRange[medium], isForBusiness[false]} & The satellite eurus 65 is a laptop designed for home use with 4 gb of memory and a medium sized hard drive \\ 
			
			\bottomrule
		\end{tabular}
		\caption{\small Sample MRs from popular NNLG datasets.}
		\label{table:traditional-MR}
	\end{footnotesize}
\end{table}

Only having simple and limited information within these MRs has several shortcomings. Model outputs are either very generic or generators have to be trained for a narrow domain and cannot be used for new domains. Thus, some recent work has focused on different methods to improve naturalness \cite{zhu2019multi-task} and promote domain transfer \cite{tran-nguyen-2018-adversarial,Wen2016MultidomainNN}.

MRs are not unique to the problem of language generation:
tasks such as dialog state tracking \cite{dstc8}, policy learning
\cite{chen-etal-2018-structured}, and task completion \cite{li-etal-2017-end}
also require the use of an MR to track context and state information relevant to
the task. MRs from these more dialog-oriented tasks are often referred to as a
``schemata.''

While dialog state tracking schemata do not necessarily include descriptions (and
generally only include names of intents, slots, and values like traditional
MRs), recent work has suggested that the use of descriptions may help with
different language tasks, such as zero-shot and transfer learning
\cite{bapna-zero}. The most recent Dialog System Technology Challenge (DSTC8)
\cite{dstc8} provides such descriptions and introduces the idea of {\it schema-guided} dialog state tracking.

Table~\ref{table:dstc8-intro-schema} shows a sample schema from DSTC8. It is
much richer and more contextually informative than traditional MRs. Each turn is
annotated with information about the current speaker, (e.g., SYSTEM, USER),
dialog act (e.g., REQUEST), slots (e.g., CUISINE), values (e.g., Mexican and
Italian), as well as the surface string utterance. When comparing this schema in
Table~\ref{table:dstc8-intro-schema} to the MRs from
Table~\ref{table:traditional-MR}, we can see that the only part of the schema
reflected in the MRs is the ACTIONS section, which explicitly describes intents,
slots, and values.

\begin{table}[h!]
	\begin{footnotesize}
		\begin{tabular}
			{@{} p{7.5cm} @{}}
			\toprule
			{\bf ACTIONS - } \\
			{ACT: } REQUEST\\ SLOT: CUISINE\\ VALUES: Mexican, Italian\\
			\textcolor{blue}{{\bf SLOT DESCRIPTIONS -} }\\
			\textcolor{blue}{CUISINE:} "Cuisine of food served in the restaurant"\\
			\textcolor{blue}{SLOT TYPE:} CUISINE: is\_categorical=true \\
			\textcolor{blue}{{\bf INTENT -}} FindRestaurants \\
			\textcolor{blue}{{\bf INTENT DESCRIPTION: }} "Find a restaurant of a particular cuisine in a city"\\
			\textcolor{blue}{{\bf SERVICE -}} Restaurants\_1 \\
			\textcolor{blue}{{\bf SERVICE DESCRIPTION:}} "A leading provider for restaurant search and reservations"\\
			\textcolor{blue}{{\bf SPEAKER -}} System \\
			\textcolor{blue}{{\bf UTTERANCE -}} "Is there a specific cuisine type you enjoy, such as Mexican, Italian, or something else?"\\
			\bottomrule
		\end{tabular}
		\caption{\small Sample schema from DSTC8. "Actions" describe a traditional MR; blue fields are newly introduced in the schema.}
		\label{table:dstc8-intro-schema}
	\end{footnotesize}
\end{table}

To our knowledge, no previous work on NNLG has attempted to generate natural
language strings from schemata using this richer and more informative
data.  In this paper, we propose the new task of \textit{Schema-guided Natural Language
	Generation}, where we take a turn-level schema as input and generate a
natural language string describing the required content, guided by the context
information provided in the schema. Following previous work on schema-guided language 
tasks, we hypothesize that descriptions in the schema will lead to better generated outputs and the possibility of zero-shot learning \cite{bapna-zero}. For example, to realize the MR {\it REQUEST(time)}, domain-specific descriptions of common slots like {\it time} can help us realize better outputs, such as {\it "What time do you want to reserve your dinner?"} in the restaurant domain, and {\it "What time do you want to see your movie?"} for movies. Similarly, we note that for dialog system developers, writing domain-specific templates for all scenarios is clearly not scalable, but providing a few domain-specific descriptions for slots/intents is much more feasible.

We focus on system-side turns from the DSTC8 dataset and, to allow our models to better generalize, we generate natural language templates, i.e., delexicalized
surface forms, such as {\it "Is there a specific cuisine type you enjoy, such as
	\$cuisine1, \$cuisine2, or something else?"} from the example schema in
Table~\ref{table:dstc8-intro-schema}.  We chose to focus on the system-side turn as currently, when building a dialog system, developers need to spend a large amount of time hand-writing prompts for each possible situation. We believe that enabling a model to automatically generate these prompts would streamline the development process and make it much faster.


Our contributions in this paper are three-fold: (1) we introduce a novel task and repurpose a dataset for {\it schema-guided NLG}, (2) we present our methods to include schema descriptions in state-of-the-art NNLG models, and (3) we demonstrate how using a schema frequently leads to better quality outputs than traditional MRs. We experiment with three different NNLG models (Sequence-to-Sequence, Conditional Variational AutoEncoders, and GPT-2 as a
Pretrained Language Model).
We show that the rich schema information frequently helps improve model performance on similarity-to-reference and semantic accuracy measures across domains, and that it promotes more diverse outputs with larger vocabularies. We also present a human evaluation demonstrating the high quality of our outputs in terms of naturalness and semantic correctness.

\section{Data}
\label{sec:data}
To create a rich dataset for NNLG, we repurpose the dataset used for the
Schema-Guided State Tracking track of
DSTC8~\cite{dstc8}.\footnote{\url{https://github.com/google-research-datasets/dstc8-schema-guided-dialogue}}
We preprocess the data to create our Schema-Guided Natural Language (SG-NLG)
dataset for training and evaluating our NNLG models.\footnote{\url{https://github.com/alexa/schema-guided-nlg}}

Since we are focused on system turns, we first drop all the user turns. The
second step in the preprocessing pipeline is to delexicalize each of the system
utterances. The original data is annotated with the spans of the
slots mentioned in each turn. We replace these mentions with the slot type
plus an increasing index prefixed by the \texttt{\$} sign, e.g.,
\texttt{\$cuisine\_1}. For example, the utterance \textit{``Is there a specific
	cuisine type you enjoy, such as Mexican, Italian, or something else?''}
becomes \textit{``Is there a specific cuisine type you enjoy, such as}
\texttt{\$cuisine\_1}\textit{,} \texttt{\$cuisine\_2} \textit{or something
	else?} 

The third step is to construct the MR corresponding to each system turn. We represent an MR as a triplet: one dialog act with exactly one slot and one value. Therefore, an MR that in the original DSTC8 dataset is represented as {\it REQUEST(cuisine = [Mexican, Italian])} becomes {\it
	REQUEST(cuisine=\$cuisine\_1), REQUEST(cuisine=\$cuisine\_2)} (see Table \ref{table:data-example}). Note that the MR has
been delexicalized in the same fashion as the utterance. Similarly, for MRs that
do not have a value, e.g., {\it REQUEST(city)}, we introduced the
\texttt{null} value resulting in {\it REQUEST(city=null)}. We also use the
\texttt{null} value to replace the slot in dialog acts that do not require one,
e.g., {\it BYE()} becomes {\it BYE(null=null)} in order to ensure that each MR is converted to a triplet.

Once we generate templates and MR pairs, we add information about the service.
In DSTC8, there are multiple {\bf services} within a single {\bf domain}, e.g., services {\it travel\_1} and {\it travel\_2} are both part of the {\it travel} domain, but have distinct schema.\footnote{We show service examples in the appendix.} DSTC8 annotates each turn with the corresponding service, so we reuse this information.
Our schema also includes user intent.\footnote{At experimentation time, the DSTC8
	test set was not annotated with user intent. Since we needed user intents for
	our task, we used DSTC8 dev as our test set. We randomly split the DSTC8
	train set into 90\% training and 10\% development.} Since only user turns are
annotated with intent information, we use the immediately preceding user turn's intent
annotation if the system turn and the user turn share the same service. If the
service is not the same, we drop the intent information, i.e., we use an empty
string as the intent (this only happens in 3.3\% of cases).

Next, we add information extracted from the schema file of the original data.
This includes service description, slot descriptions (one description for each
slot in the MR), and intent descriptions. These descriptions are very short English sentences (on average 9.8, 5.9 and 8.3 words for services, slots and intents).	 Lastly, we add to each triplet a
sentence describing, in plain English, the meaning of the MR. These description
are not directly available in DSTC8 but are procedurally generated by a set of
rules.\footnote{We have a single rule for each act type; $10$ in total.} For
example, the MR {\it CONFIRM(city=\$city\_1)} is \textit{``Please confirm that
	the [city] is [\$city\_1].''} The intuition behind these \textit{natural
	language MRs} is to provide a more semantically informative representation of
the dialog acts, slots and values.


Table~\ref{table:dataset_dim} shows the SG-NLG dataset statistics. In
summary, SG-NLG is composed of nearly 4K MRs and over 140K templates. On
average, every MR has 58 templates associated with it, but there is a large
variance. There is one MR associated with over 1.7K
templates ({\it CONFIRM(restaurant\_name, city, time, party\_size, date)}) and many MRs with only one template. 

\begin{table}[h!]
	\begin{footnotesize}
		\begin{tabular}
			{@{} p{7.5cm} @{}}
			\sc DSTC8 (original) \\
			\toprule
			{\bf ACTIONS - } \\
			{ACT: } REQUEST\\ SLOT: CUISINE\\ VALUES: Mexican, Italian \\
			{\bf UTTERANCE -} "Is there a specific cuisine type you enjoy, such as Mexican, Italian, or something else?"\\\hline
			
			\addlinespace[2ex]
			\sc SG-NLG (pre-processed)\\
			\toprule
			{\bf MR}=[REQUEST(cuisine=\$cuisine1), \\
			$\:\:\:\:\:\:\:\:\:\:\:$REQUEST(cuisine=\$cuisine2)] \\
			{\bf UTTERANCE -} "Is there a specific cuisine type you enjoy, such as \$cuisine1, \$cuisine2, or something else?"\\
			\bottomrule
		\end{tabular}

		\caption{\small Data preprocessing and delexicalization.}
		\label{table:data-example}
	\end{footnotesize}
\end{table}

\begin{table}[h!]
	\begin{footnotesize}
		\begin{tabular}{l r r r}
			& \textbf{Train} & \textbf{Dev} & \textbf{Test} \\\hline
			Templates & 110595 & 14863 & 20022\\
			Meaning Representations & 1903 & 1314 & 749\\
			Services & 26 & 26 & 17 \\
			Domains & 16 & 16 & 16 \\
		\end{tabular}
	\end{footnotesize}
	\centering \caption{\label{table:dataset_dim} {\small SG-NLG dataset
			statistics.}}
\end{table}

\section{Models}
\label{sec:models}

\subsection{Feature Encoding}\label{sec:features}
We categorize the features from schemata into two different types. The first type
is \textbf{symbolic features}. Symbolic features are encoded using a
word embedding layer. They typically consist of single tokens, e.g., service
names or dialog acts, and frequently resemble variable names (e.g., \texttt{restaurant} and
\texttt{restaurant\_name}). The second type of features is
\textbf{natural language features}. These features are typically sentences,
e.g., service/slot descriptions or the natural language MR, that we encode using BERT \cite{devlin2018bert} 
to derive a single semantic embedding tensor.

\begin{figure}[h!]
	\includegraphics[width=\linewidth]{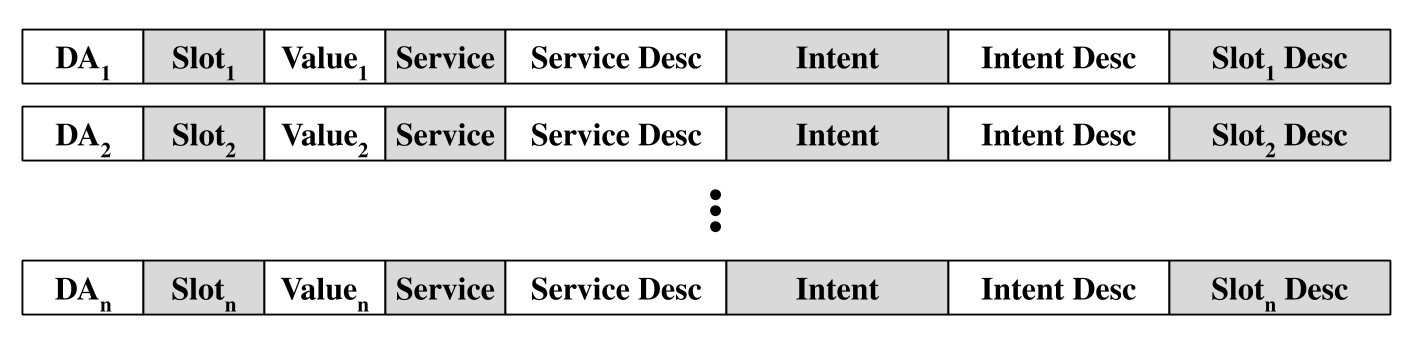}
	\caption{Flat-encoding strategy.}\label{fig:embedding}
\end{figure}

To represent the full schema, we adopt a flat-encoding strategy. The first 
part of each schema is the MR, which we define as a sequence of dialog act, slot, and value tuples.
At each timestep, we encode a three-part sequence: (1) a new act, slot, and value tuple from the MR, 
(2) the embeddings of all schema-level features (i.e., services, intents, and their descriptions), and (3) the embedding of the current slot description (see Figure \ref{fig:embedding}). Finally, we append the encoded natural language MR.

\subsection{Sequence-to-Sequence}
Our first model is a Seq2Seq model with attention, copy, and constrained decoding (see the full model diagram in the appendix).
We implement the attention from \citet{luong2015effective}:
$$a_t = \textit{softmax}(align(h_t, s_t))$$ where $align$ is a function that computes the alignment score of the hidden state of the encoder $h_t$ and the decoder hidden state, $s_t$. The goal of this layer is to attend to the more salient input features.


The copy mechanism we add is based on pointer-generator
networks~\cite{see2017get}. At each decoding
step $t$ we compute a probability $p_{gen}$:
$$p_{gen} =\sigma(w_h^T h_t^* + w_s^Ts_t + w_x^Tx_t + b_{ptr})$$
where $w_h$, $w_s$, and $w_x$ are a learnable weights matrix; $h_t^*$ is a context vector computed by combining the encoder hidden state and the attention weights, $s_t$ is the decoder hidden state, $x_t$ the decoder  input, and $b_{ptr}$ is a bias term. The probability $p_{gen}$ is then used to determine the next word $w$ generated:
$$P(w) = p_{gen}P_{vocab}(w) + (1-p_{gen})\sum_{i}a_i^t$$
Thus $p_{gen}$ behaves like a switch to decide whether to generate from the vocab or copy from the input. The goal of the copy mechanism is to enable the generation of special symbols such as {\tt \$cuisine\_1} that are specific to the service. 

\subsection{Conditional Variational Auto-Encoder}

The Conditional Variational Auto-Encoder (CVAE)~\cite{hu2017toward} is an
extension of the VAE models, where an additional vector $c$ is attached to the
last hidden state of the encoder $z$ as the initial hidden state of the decoder.
The vector $c$ is used to control the semantic meaning of the output to align
with the desired MR. We use the encoded feature vector described in
Section~\ref{sec:features} as $c$.
The model objective is the same as VAE, which is the sum of reconstruction loss and Kullback–Leibler divergence loss. At training time, $z$ is the encoded input sentence. At prediction time, $z$ is sampled from a Gaussian prior learned at training time. We also adapt the attention mechanism for CVAE by adding an additional matrix $W_{e}$ to compute the alignment score,
$$align(h_t, \tilde{s_t}) = W(W_e * h_t + \tilde{s_t}))$$
where $\tilde{s_t}$ is the decoder hidden state.

For Seq2Seq/CVAE, we use constrained decoding to prune out
candidate outputs with slot repetitions. We use a beam to keep track of slots that have already been generated and set the probability of a new
candidate node to zero if slots are repeated.

\begin{table*}[t!]
	\begin{footnotesize}
		\begin{tabular}{p{1cm}|p{14cm}} \hline 
			\rowcolor{lightgray}
			\multicolumn{2}{l}{{\bf [Schema 1] } \it \underline{ACTIONS (MR):} INFORM(price-per-night= \$price-per-night1), NOTIFY-SUCCESS(null=null)} \\
			\rowcolor{lightgray}
			\multicolumn{2}{l}{\it \underline{Slot Desc:} price-per-night: "price per night for the stay"} \\
			\rowcolor{lightgray}
			\multicolumn{2}{l}{\it \underline{Service:} hotels-4 $\:\:\:\:\:\:\:\:\:\:$  \underline{Service Desc:} "Accommodation searching and booking portal"}\\
			\rowcolor{lightgray}
			\multicolumn{2}{l}{\it \underline{Intent:} ReserveHotel $\:\:\:\:$ \underline{Intent Desc:} "Reserve rooms at a selected place for given dates."} \\
			\rowcolor{lightgray}
			\multicolumn{2}{l}{\it \underline{Natural Language MR:} the [price per night] is [\$price-per-night1]. the request succeeded.} \\
			\rowcolor{lightgray}
			\hline
			\rowcolor{LightCyan}
			\it Ref & \it  \$price-per-night1 a night \\
			Seq2Seq & your reservation is booked and the total cost is \$price-per-night1 .\\
			CVAE & your reservation has been made . the total cost is \$price-per-night1 per night .\\
			GPT2 & your reservation was successful! the cost of the room is
			\$price-per-night1 per night. \\ \hline \hline
			
			\rowcolor{lightgray}
			\multicolumn{2}{l}{{\bf [Schema 2] }\it \underline{ACTIONS (MR):} OFFER(movie-name= \$movie-name1),
				OFFER(movie-name= \$movie-name2)}\\
			\rowcolor{lightgray}
			\multicolumn{2}{l}{\it OFFER(movie-name= \$movie-name3), INFORM(count=\$count1)} \\
			\rowcolor{lightgray}
			\multicolumn{2}{l}{\it \underline{Slot Desc:} movie-name: "name of the movie", count: "the number of items that satisfy the user's request"} \\
			\rowcolor{lightgray}
			\multicolumn{2}{l}{\it \underline{Service:} media-2 $\:\:\:\:\:\:\:$  \underline{Service Desc:} "The widest selection and lowest prices for movie rentals"}\\
			\rowcolor{lightgray}
			\multicolumn{2}{l}{\it \underline{Intent:} FindMovies $\:\:\:\:$ \underline{Intent Desc:} "Find movies to watch by genre and, optionally, director or actors"} \\
			\rowcolor{lightgray}
			\multicolumn{2}{l}{\it \underline{Natural Language MR:} there is [\$movie-name2] for [movie name]. }\\
			\rowcolor{lightgray}
			\multicolumn{2}{l}{\it $\:\:\:\:$there is [\$movie-name3] for [movie name]. there is [\$movie-name1] for [movie name]. the [count] is [\$count1].} \\\hline
			\rowcolor{LightCyan}
			\it Ref & \it  \$count1 movies \$movie-name1 \$movie-name2 and \$movie-name3 \\
			Seq2Seq & i found \$count1 movies . how about \$movie-name2 ?\\
			CVAE & i found \$count1 movies you might like . how about \$movie-name1 , \$movie-name2 or \$movie-name3 ? \\
			GPT2 & sure. hey! i've got \$count1 movies. do you wish to watch any
			of these movies - \$movie-name1, \$movie-name2 or \$movie-name3? \\ \hline
			\hline
			
			\rowcolor{lightgray}
			\multicolumn{2}{l}{{\bf [Schema 3] }\it \underline{ACTIONS (MR):} REQUEST(transfer-amount=null), REQUEST(recipient-name=null)}\\
			\rowcolor{lightgray}
			\multicolumn{2}{l}{\it \underline{Slot Desc:} transfer-amount: "the amount of money to transfer", recipient-name: "the name of the recipient to transfer [...]"} \\
			\rowcolor{lightgray}
			\multicolumn{2}{l}{\it \underline{Service:} banks-2 $\:\:\:\:\:\:\:\:\:\:\:\;$  \underline{Service Desc:} "Service to manage your bank accounts and finances"}\\
			\rowcolor{lightgray}
			\multicolumn{2}{l}{\it \underline{Intent:} TransferMoney $\:\:\:\:$ \underline{Intent Desc:} "Transfer money to another user"} \\
			\rowcolor{lightgray}
			\multicolumn{2}{l}{\it \underline{Natural Language MR:} "what [transfer amount] do you want? what [recipient name] do you want?"} \\\hline
			\rowcolor{LightCyan}
			\it Ref & \it  amount? recipient? \\
			Seq2Seq & what type of ride do you want to transfer ?\\
			CVAE & what is your destination ? how much do you want to send the money to ?\\
			GPT2 & sure, i can help you with that. how much would you like to
			transfer from? who would you like to send it to? \\ \hline
			
		\end{tabular}
	\end{footnotesize}
	\vspace{-.1in}
	\centering \caption{\label{table:output-examples} {\small Example model
			outputs. All models are correct for Schema 1. Seq2Seq model dropped two slots
			for Schema 2. Schema 3 shows incorrect outputs for Seq2Seq and
			CVAE.}}
\end{table*} 

\subsection{Pretrained Language Model: GPT-2}
We also experiment with a pretrained language model, specifically
GPT-2~\cite{radford2019language}.\footnote{We use GPT-2 small from HuggingFace
	Transformers (\url{https://github.com/huggingface/transformers})} Since GPT-2 is
trained on purely natural language strings, we first combine the symbolic and
natural language features into flat natural language strings, similar to
previous work by \citet{budzianowski-vulic-2019-hello}. We fine-tune the GPT-2
model using these natural language inputs with the target template.\footnote{We
	train with special beginning of sequence, end of sequence, and separator tokens
	such that each training instance is: ``{\tt [BOS]} schema-tokens {\tt[SEP]}
	target-tokens {\tt[EOS]}.''} At prediction time, given the schema tokens as input, we use our fine-tuned GPT-2 model with a language model head to generate an output sequence (until we hit an end-of-sequence token). We adopt top-k sampling at each decoding step. 

\section{Evaluation}
\label{sec:evaluation}

For each of our three models, we generate a single output for each test
instance. Table~\ref{table:output-examples} shows example model outputs. 

\subsection{Evaluation Metrics}
We focus on three distinct metric types: similarity to references, semantic accuracy, and diversity.

{\bf Similarity to references.} As a measure of how closely our outputs
match the corresponding test references, we use BLEU (n-gram precision with
brevity penalty) \cite{eval_bleu} and METEOR (n-gram precision and recall, with synonyms) \cite{eval_meteor}. We compute corpus-level BLEU for the full set of outputs and matching references. For METEOR, we compute per-output metrics and average across all instances.\footnote{We use NLTK for BLEU4/METEOR \cite{nltk-book}.} We include these metrics in our evaluation primarily for completeness and supplement them with a human evaluation, since it is widely agreed that lexical overlap-based metrics are weak measures of quality \cite{novikovanewnlg,belz-reiter-2006-comparing,bangalore-etal-2000-evaluation}.

\begin{table*}[t!]
	\begin{footnotesize}
		\begin{tabular}{p{1cm}|p{1.3cm}|p{1cm}p{1.1cm}|p{1cm}p{1cm}|p{0.9cm}p{0.9cm}p{1cm}p{1cm}p{1cm}}
			& & \multicolumn{2}{|c|}{\bf Similarity to Refs}  & \multicolumn{2}{|c|}{\bf Semantics}  & \multicolumn{5}{|c}{\bf Diversity}  \\ \hline
			&        &        BLEU Corpus & METEOR Avg   & SER Avg $\downarrow$ & Slot Match Rate & Vocab1 (Gold: 2.5k) & Vocab2 (Gold: 20k) & Distinct1 (Gold: 0.01) & Distinct2 (Gold: 0.1) & Novelty \\ \hline
			Seq2Seq  &  MR           & 0.4059  & 0.5254                      &  \bf 0.1602               &  \bf 0.7530              & 253                     & 614                    & 0.0398  & 0.1093 & \bf  0.5741 \\
			& Schema & \bf 0.4174  &  \bf 0.5580                      & 0.2062               & 0.7009              &  \bf 275                     &  \bf 699                    &  \bf 0.0445  &  \bf 0.1288 & 0.5674 \\ \hline
			CVAE     & MR           & 0.4282  & 0.5595                      & 0.2469               & 0.6622              & 292                     & 727                    & 0.0406  & 0.1128 & 0.5434 \\
			& Schema & \bf 0.4299  &\bf 0.5852                      &\bf 0.2407               &\bf 0.6983              &\bf 327                     &\bf 924                    &\bf 0.0445  &\bf 0.1401 &\bf 0.6142 \\ \hline
			GPT2     &  MR           & 0.3551  & 0.5640                      & 0.1929               & 0.8331              & 648                     & 2491                   & 0.0818  & 0.3471 & 0.5808 \\
			&Schema &  \bf 0.4030  &  \bf 0.6129                      &  \bf 0.1810               &  \bf 0.8558              &  \bf 678                     &  \bf 2659                   &  \bf 0.0868  &  \bf 0.3767 &  \bf 0.5955 \\
		\end{tabular}
	\end{footnotesize}
	\vspace{-.1in}
	\centering \caption{\label{table:results-mr-vs-schema} {\small Automatic
			evaluation metrics comparing traditional MR vs.\@ rich schema.
			Higher is better for all metrics except SER.}}
\end{table*}

{\bf Semantic accuracy.} We compute the slot error rate (SER) for each
model output as compared to the corresponding MR by finding the total number of
deletions, repetitions, and hallucinations over the total number of slots for
that instance (the lower the better).\footnote{Although
	\citet{Wen15} compute SER using only deletions and repetitions, we include
	hallucinations to capture errors more accurately.} It is important to note that we only consider slots that have explicit values (e.g., {\it MR: INFORM date=\$date1}) for our automatic SER computations. We are investigating methods to compute SER over implicit slots (e.g., {\it MR: REQUEST party\_size=null}) as future work, since it is non-trivial to compute due to the various ways an implicit slot might be expressed in a generated template (e.g., {\it "How many people are in your party?"}, or {\it "What is the size of your group?"}). We also compute "slot match rate", that is the ratio of generated outputs that contain exactly the same explicit slots as the matching test MR. 

{\bf Diversity.} We measure diversity based on vocabulary, distinct-$N$
(the ratio between distinct $n$-grams over total $n$-grams)
\cite{li-etal-2016-diversity} and novelty (the ratio of unique generated
utterances in test versus references in train).\footnote{To avoid inflating
	novelty metrics, we normalize our template values. (e.g., {\it "Table is reserved for \$date1."} is normalized to {\it "Table is reserved for \$date."} for any {\it \$dateN} value).}

\subsection{Traditional MR vs.\@ Rich Schema}
Table \ref{table:results-mr-vs-schema} compares model performance when trained using only the traditional MR versus using the full schema (better result for each model in bold). 

{\bf Model comparisons.} To get a general sense of model performance, we first compare results across models. From the table, we see that Seq2Seq and CVAE have higher BLEU compared to GPT2 (for both MR and Schema), but that GPT2 has a higher METEOR. This indicates that GPT2 is more frequently able to generate outputs that are semantically similar to references, but that might not be exact lexical matches (e.g., substituting "film" for "movie") since GPT2 is a pretrained model. Similarly, GPT2 has a significantly higher vocabulary and diversity than both Seq2Seq and CVAE.

{\bf MR vs. Schema.} Next, we compare the performance of each model when trained using MR versus Schema. For all models, we see an improvement in similarity metrics (BLEU/METEOR) when training on the full schema. Similarly, in terms of diversity, we see increases in vocabulary for
all models, as well as increases in distinct-$N$ and novelty (with the exception of Seq2Seq novelty, which drops slightly). 

In terms of semantic accuracy, we see an improvement in both SER and Slot Match
Rate for both CVAE and GPT2. For Seq2Seq, however, we see that the
model performs better on semantics when training on only the MR. To investigate,
we look at a breakdown of the kinds of errors made. We find that Seq2Seq/CVAE
only suffer from deletions, but GPT2 also produces repetitions and
hallucinations (a common problem with pretrained language models); however, training using the schema reduces the number of these
mistakes enough to result in an SER improvement for GPT2 (see the appendix for details).

\begin{table}[b!]
	\begin{small}
		\begin{tabular}{p{0.5cm}|p{0.7cm}p{0.7cm}|p{0.7cm}p{0.7cm}|p{0.7cm}p{0.7cm}}\toprule
			& \multicolumn{2}{c|}{\sc Seq2Seq} & \multicolumn{2}{c|}{\sc CVAE} & \multicolumn{2}{c}{\sc GPT2} \\\hline	
			& \bf BLEU & \bf SER$\downarrow$ & \bf BLEU 	& \bf SER$\downarrow$ & \bf BLEU & \bf SER$\downarrow$\\\hline
			\addlinespace[2ex] 
			\multicolumn{7}{l}{\bf Seen} \\\hline
			MR & 0.51 & \bf 0.07 &  0.56 & 0.12 & 0.46 & 0.05 \\
			Sch & \bf 0.57 &  0.12 & \bf 0.61 & \bf 0.09 & \bf 0.51 & \bf 0.04 \\\hline 
			\addlinespace[2ex]
			\multicolumn{7}{l}{\bf Partially-Unseen} \\\hline
			MR & \bf 0.38 & \bf 0.23 & \bf 0.38 & \bf 0.34 & 0.33 & 0.31 \\
			Sch & \bf 0.38 & 0.28 & 0.33 & 0.37 & \bf 0.38 & \bf 0.29 \\\hline
			\addlinespace[2ex]
			\multicolumn{7}{l}{\bf Fully-Unseen} \\\hline
			MR & 0.34 &\bf  0.27 & 0.34 & \bf 0.27 & 0.16 & \bf 0.48 \\
			Sch & \bf 0.36 & \bf 0.27 & \bf 0.45 & \bf 0.27 & \bf 0.22 & 0.58 \\
			\bottomrule
		\end{tabular}
	\end{small}
	\vspace{-.1in}
	\centering \caption{\label{table:results-average-mr-vs-schema} {\small
			Average BLEU and SER by service splits.}}
\end{table}

\begin{table*}[t!]
	\begin{footnotesize}
		
		\begin{center}
			\begin{subtable}[t]{\textwidth}
				\begin{tabular}{p{2cm}P{1.7cm}|P{1.5cm}P{1.5cm}|P{1.5cm}P{1.5cm}|P{1.5cm}P{1.3cm}}
					&              & \multicolumn{2}{|c|}{\sc Seq2Seq}     & \multicolumn{2}{c|}{\sc CVAE}        & \multicolumn{2}{c}{\sc GPT2}        \\\hline
					\bf Service         & \bf \% Test Refs & \bf BLEU  & \bf SER$\downarrow$  & \bf BLEU  & \bf SER$\downarrow$  & \bf BLEU  & \bf SER$\downarrow$  \\\hline
					{\tt\footnotesize events\_1}      & 19\%         & 0.6168   &\bf 0.0490  & 0.6126     & \bf 0.0294  & 0.4682   & 0.0588  \\
					{\tt\footnotesize rentalcars\_1}  & 18\%         &  \bf 0.7486  & 0.1500  &   \bf 0.6645   & \it 0.1125  &\bf 0.6173   & \it 0.1000  \\
					{\tt\footnotesize buses\_1}       & 15\%         & \it 0.3831  &\it 0.1542  & \it 0.5035    & 0.1000  & \it 0.4016   & \bf 0.0167  \\
				\end{tabular}
				\caption{\label{table:seen} {\small Seen services.}}
			\end{subtable}
		\end{center}
		\vfill
		
		\begin{center}
			\begin{subtable}[t]{\textwidth}
				\begin{tabular}{p{2cm}P{1.7cm}|P{1.5cm}P{1.5cm}|P{1.5cm}P{1.5cm}|P{1.5cm}P{1.3cm}}
					&              & \multicolumn{2}{|c|}{\sc Seq2Seq}     & \multicolumn{2}{c|}{\sc CVAE}        & \multicolumn{2}{c}{\sc GPT2}        \\\hline
					\bf Service         & \bf \% Test Refs & \bf BLEU  & \bf SER$\downarrow$  & \bf BLEU  & \bf SER$\downarrow$  & \bf BLEU  & \bf SER$\downarrow$  \\\hline
					{\tt\footnotesize restaurants\_2} & 24\%         &\it 0.2466   &  \bf 0.2098  &  \it 0.2126   &\bf 0.3501  &  \it 0.2297   &  \bf 0.0527  \\
					{\tt\footnotesize flights\_3}     & 18\%         &   0.3193  &  \it 0.4579  & \bf 0.3481   &  \it 0.5000  & 0.3008    & \it 0.7368  \\
					{\tt\footnotesize services\_4}    & 18\%         &  \bf 0.5791  & 0.2197  &  0.3288   & 0.4013  &  \bf 0.5760 & 0.0851  \\
					
				\end{tabular}
				\caption{\label{table:partially-unseen} {\small Partially-unseen
						services.}}
			\end{subtable}
		\end{center}
		\vfill
		\begin{center}
			\begin{subtable}[t]{\textwidth}
				\begin{tabular}{p{2cm}P{1.7cm}|P{1.5cm}P{1.5cm}|P{1.5cm}P{1.5cm}|P{1.5cm}P{1.3cm}}
					&              & \multicolumn{2}{|c|}{\sc Seq2Seq}     & \multicolumn{2}{c|}{\sc CVAE}        & \multicolumn{2}{c}{\sc GPT2}        \\\hline
					\bf Service         & \bf \% Test Refs & \bf BLEU  & \bf SER$\downarrow$  & \bf BLEU  & \bf SER$\downarrow$  & \bf BLEU  & \bf SER$\downarrow$  \\\hline
					{\tt\footnotesize alarm\_1}       & 100\%        & 0.3586  & 0.2667  & 0.4495 & 0.2667  & 0.2217    & 0.5833 \\
				\end{tabular}
				\caption{\label{table:fully-unseen} {\small Fully-unseen
						services.}}
			\end{subtable}
		\end{center}
	\end{footnotesize}
	\vspace{-.1in}
	\caption{\label{table:results-domains} {\small Automatic evaluation metrics
			across seen, partially-unseen, and fully-unseen services when training with schema.}}
\end{table*}

\subsection{Seen vs.\@ Unseen Services}
Next, we are interested to see how our models perform on specific services in the SG-NLG dataset. Recall that the original dataset consists of a set of services that can be grouped into domains: e.g., services {\tt restaurant\_1} and {\tt restaurant\_2} are both under the {\tt restaurant} domain. Based on this, we segment our test set into three parts, by service: {\it seen}, or services that have been seen in training, {\it partially-unseen}, or services that are unseen in training but are part of domains that have been seen, and {\it fully-unseen} where both the service and domain are unseen.\footnote{We show distribution plots by service in the appendix.}

{\bf MR vs. Schema.} To better understand how the models do on average across all services, we show average BLEU/SER scores in
Table~\ref{table:results-average-mr-vs-schema}.\footnote{Scores are weighted by the percentage of test references per service in each split, e.g. {\tt events\_1} in {\it seen} makes up 19\% of the {\it seen} test references, thus its scores are weighted by that factor.} Once again, we compare performance
between training on the MR vs.\@ the schema. On average, we see that
for the seen and fully-unseen partitions, training with the schema is better
across almost all metrics (sometimes showing no differences for SER for fully unseen). For partially-unseen, we see that CVAE performs
better when training on only the MR; however, when averaging across the full test in Table~\ref{table:results-mr-vs-schema}, we see an improvement with schema. 

We see naturally higher BLEU and lower SER for seen vs.\@ both partially-unseen and fully-unseen across all models. Surprisingly, we see higher schema BLEU for CVAE on fully-unseen as compared to partially-unseen, but we note that there is a very small fully-unseen sample size (only 10 test MRs). We also note that GPT2 has high SER for the fully-unseen domain; upon inspection, we see slot hallucination from GPT2 within {\tt alarm\_1}, while Seq2Seq/CVAE never hallucinate.

{\bf Seen vs.\@ Unseen.} Table \ref{table:results-domains} shows model performance in terms of BLEU and SER. We sort services by how many references we have for them in test; {\tt events\_1} for example constitutes 19\% of the test references.  To focus our discussion here, we show only the top-3 services in terms of percentage of test references.\footnote{We show results for all services in the appendix.} For fully-unseen we show the only available service ({\tt alarm\_1}). We show the best scores in bold and the worst scores in italic. 

For seen services (Figure \ref{table:seen}), we see the highest BLEU scores for all models on the {\tt rentalcars\_1}. We note that SER is consistently low across all models, with the worst SER for the top-3 services at 0.15 (the worst SER across all of seen is 0.23 as shown in the appendix). 

For partially-unseen services (Figure \ref{table:partially-unseen}), we see the best SER on {\tt restaurants\_2} (but comparatively lower BLEU scores). The {\tt services\_4} domain shows the highest BLEU scores for Seq2Seq and GPT2, with low SER. We note that {\tt flights\_3} has the worst SER for all models. Upon investigation, we find slot description discrepancies: e.g., slot {\tt origin\_airport\_name} has slot description {\it "Number of the airport flying out from"}. This highlights how models may be highly sensitive to nuances in the schema information, warranting further analysis in the future.

\subsection{Human Evaluation}
To supplement our automatic metric evaluations which show some the benefits of schema-based generation, we conduct an annotation study to evaluate our schema-guided output quality. We randomly sample 50 MRs from our test set, and collect 3 judgments per output for each model as well as a reference (randomly shuffled).\footnote{We have a pool of 6 annotators that are highly-skilled at evaluating language tasks and were not involved in any other parts of the project.}

We ask the annotators to give a binary rating for each output across 3 dimensions: {\it grammar}, {\it naturalness}, and {\it semantics} (as compared to the input MR). We also get an "overall" rating for each template on a 1 (poor) to 5 (excellent) Likert scale.\footnote{To make annotation more intuitive, we automatically lexicalize slots with values from the schema (although this may add noise), e.g., "The date is \$date1" $\rightarrow$ "The date is [March 1st]." We use the same values for all templates for consistency.}

Table~\ref{table:results-human-eval} summarizes the results of the study. For grammar, naturalness, and semantics, we show the ratio of how frequently a given model or reference output is marked as correct over all outputs for that model. For the "overall" rating, we average the 3 ratings given by the annotators for each instance, and present an average across all MRs (out of 5).

\begin{table}[h!]
	\begin{small}
		\begin{tabular}{p{1cm}p{1.2cm}p{1.2cm}p{1.2cm}p{1.2cm}}\toprule
			&  Grammar (\%) &  Naturalness (\%)   & Semantics (\%) & Overall (out of 5)   \\ \hline
			Reference & 0.95 & 0.67 & 0.91 & 3.97 \\\hline
			Seq2Seq & 0.82 & 0.58 & 0.37 & 2.72 \\
			CVAE & \bf 0.89 & \bf 0.73 & 0.44 & 3.01 \\
			GPT2 & 0.80 & 0.61 & \bf 0.70 & \bf 3.61\\
			\bottomrule
		\end{tabular}
	\end{small}
	\vspace{-.1in}
	\centering \caption{\label{table:results-human-eval} {\small Average human
			evaluation scores for different quality dimensions.}}
\end{table}

From the table, we see that the CVAE model has the
highest score in terms of both grammar and naturalness. Moreover, CVAE
also achieves a score higher than the reference in terms of naturalness. A possible explanation explanation for this behavior is that the quality of the reference is subjective, and not always an ideal "gold-standard". In terms
of semantics, we see that GPT-2 has the highest ratings of all models. Most
interestingly, we see that CVAE has a significantly lower semantic rating,
although it is the winner on grammar and naturalness, indicating that while CVAE
outputs may be fluent, they frequently do not actually express the required
content (see Schema 3 in Table~\ref{table:output-examples}). This finding is
also consistent with our SER calculations from
Table~\ref{table:results-mr-vs-schema}, where we see that CVAE has the highest SER.\footnote{We compute Fleiss Kappa scores for each dimension, finding near-perfect agreement for semantics (0.87), substantial agreement for grammar (0.76), and moderate agreement for naturalness (0.58) and overall (0.47).}

In terms of overall score, we see that GPT-2 has the highest rating of all three models, and is most frequently comparable to the ratings for the references. This can be attributed to its higher semantic accuracy, combined with good (even if not the highest) ratings on grammar and naturalness.

\section{Related Work}
\label{sec:related-work}
Most work on NNLG uses a simple MR that consists of slots and value tokens that only describe information that should be realized, without including contextual information to guide the generator as we do; although some work has described how this could be useful \cite{walker-etal-2018-exploring}. WebNLG \cite{webnlg_challenge} includes structured triples from Wikipedia which may constitute slightly richer MRs, but are not contextualized. Oraby et al. \shortcite{oraby-etal-2019-curate} generate rich MRs that contain syntactic and stylistic information for generating descriptive restaurant reviews, but do not add in any contextual information that does not need to be included in the output realization. Table-to-text generation using ROTOWIRE (NBA players and stats) also includes richer information, but it is also not contextualized \cite{wiseman2017challenges, Gong2019TabletoTextGW}.

Other previous work has attempted to address domain transfer in NLG. Dethlefs et al. \shortcite{Dethlefs2017DomainTF} use an abstract meaning representation (AMR) as a way to share common semantic information across domains. Wen et al. \shortcite{Wen2016MultidomainNN} use a "data counterfeiting" method to generate synthetic data from existing domains to train models on unseen domains, then fine-tune on a small set of in-domain utterances. Tran et al. \shortcite{tran-nguyen-2018-adversarial} also train models on a source domain dataset, then fine-tune on a small sample of target domain utterances for domain adaptation. Rather than fine-tuning models for new domains, our data-driven approach allows us to learn domain information directly from the data schema.

\section{Conclusions}
\label{sec:conclusions}
In this paper, we present the novel task of \textit{Schema-Guided NLG}. We demonstrate how we are able to generate templates (i.e., delexicalized system prompts) across different domains using three state-of-the-art models, informed by a rich schema of information including intent descriptions, slot descriptions and domain information. We present our novel SG-NLG dataset, which we construct by re-purposing a dataset from the dialog state tracking community.  

In our evaluation, we demonstrate how training using our rich schema frequently improves the overall quality of generated prompts. This is true for different similarity metrics (up to 0.43 BLEU and 0.61 METEOR) that we recognize are weak measures of quality but, more importantly, for semantic metrics (as low as 0.18 average SER), and even for diversity (up to 2.6K bigram vocabulary). Moreover, this holds true on both seen and unseen domains in many different settings. We conduct a human evaluation as a more accurate quality assessment, and show how our outputs are rated up to 3.61 out of 5 overall (as compared to 3.97 for references). We observe that different models have different strengths: Seq2Seq and CVAE have higher BLEU reference similarity scores, while GPT2 is significantly more diverse and is scored highest overall in human evaluation.


For future work, we are interested in exploring how schema-guided NLG can be used in dialog system contexts, where only outputs that have no slot errors and high overall fluency should be selected as  responses. We are also interested in improving both the semantic correctness and fluency of our model outputs by introducing improved methods for constrained decoding and language model integration. Additionally, we plan to develop more accurate automatic measures of quality, as well as more fine-grained control of domain transfer.

\section*{Acknowledgments}
\label{sec:acknowledgment}
The authors would like to thank Sofia Scharfenberg, Jasmin Rehm, and the rest of the Alexa Data Services Rapid Machine Learning Prototyping team for all of their help with preparing and performing the human evaluation study.

\bibliography{anthology,acl2020}
\bibliographystyle{acl_natbib}

\newpage
\appendix
\section*{Appendix}
\label{sec:appendix}
\section{Service and Slot Descriptions}

\begin{table}[!htbp]
	\begin{footnotesize}
		\centering
		\caption{Services, slots and their descriptions. In boldface the service names, in verbatim the slots.}
		\begin{tabular}{p{0.2\textwidth} p{0.68\textwidth}}
			\toprule
			\textbf{Events\_1} &  The comprehensive portal to find and reserve seats at events near you\\
			\texttt{category} & Type of event\\
			\texttt{time} & Time when the event is scheduled to start\\
			\midrule
			\textbf{Events\_2} &  Get tickets for the coolest concerts and sports in your area\\
			\texttt{date} & Date of event\\
			\texttt{time} & Starting time for event\\
			\midrule
			\textbf{Media\_1} &  A leading provider of movies for searching and watching on-demand\\
			\texttt{title} & Title of the movie\\
			\texttt{genre} & Genre of the movie\\
			\bottomrule
		\end{tabular}
	\end{footnotesize}
\end{table}


%

\section{Details of SER Errors}
All of the errors made by Seq2Seq and CVAE are deletion errors (constrained decoding prevents repetitions/hallucinations). While using schema leads to more deletions in GPT2, it reduces repetitions and hallucinations, leading to better SER.

\begin{table}[!h]
	\begin{small}
		\begin{tabular}{p{1cm}|p{0.8cm}|p{0.9cm}|p{0.9cm}p{0.9cm}p{0.7cm}}\toprule
			&        &        SER$\downarrow$ & Delete     & Repeat & Halluc.\\ \hline
			Seq2Seq & MR & \bf 0.1602     & \bf 0.1602            & 0      & 0      \\
			& Schema         & 0.2062     & 0.2062            & 0      & 0      \\\hline
			CVAE    &  MR & 0.2469     & 0.2469            & 0      & 0      \\
			& Schema         &  \bf 0.2407     &  \bf  0.2407            & 0      & 0     \\ \hline
			GPT2    & MR & 0.1929     & \bf 0.0791            & 0.0037 & 0.1101 \\
			& Schema         & \bf 0.1810     & 0.0850            & \bf 0.0020 & \bf 0.0940 \\
			\bottomrule
		\end{tabular}
	\end{small}
	\vspace{-.1in}
	\centering \caption{\label{table:results-ser} Detailed analysis of slot errors.}
\end{table}

\section{Seen vs. Unseen Domains}
\subsection{Data Distribution Plots}

\begin{figure}[h!]
	\centering
	\begin{subfigure}[h]{\textwidth}
		\includegraphics[width=\textwidth]{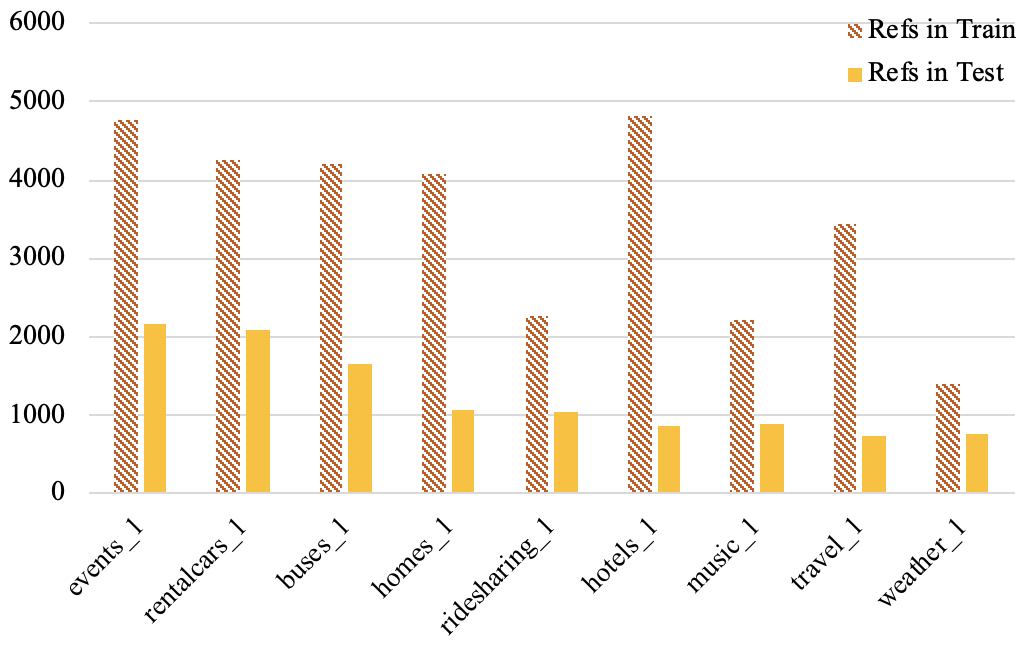}\caption{Distribution of refs in seen services.\\}\label{fig:seen}
	\end{subfigure}
	\begin{subfigure}[h]{\textwidth}
		\includegraphics[width=\textwidth]{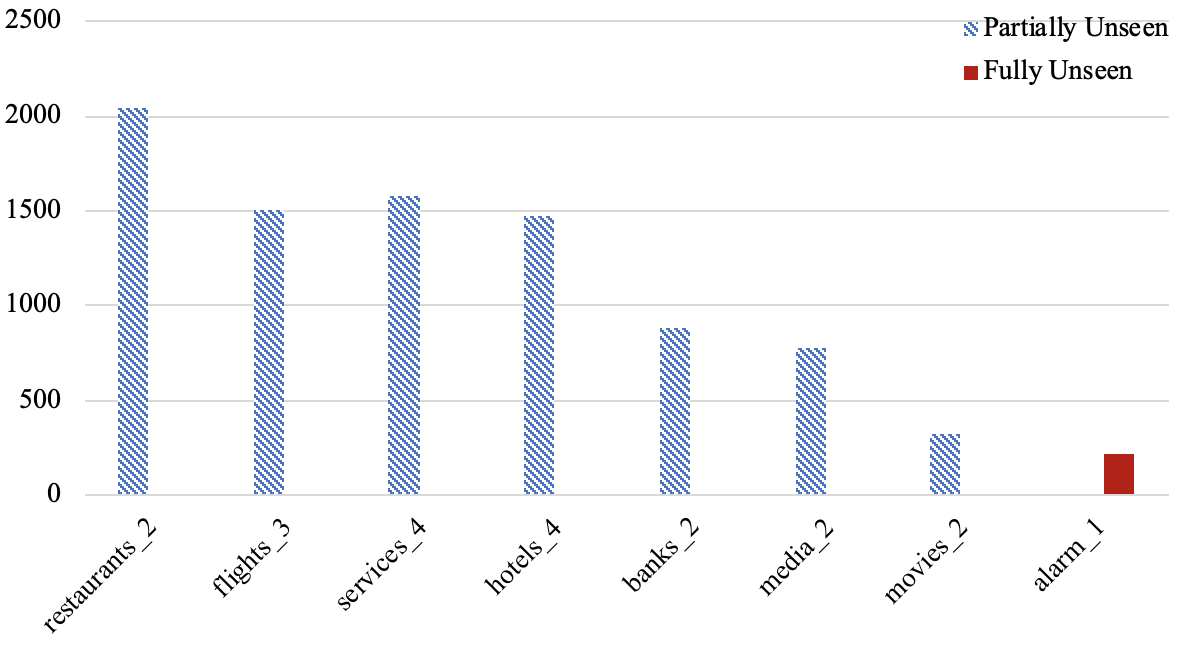}\caption{Distribution of refs in partially/fully unseen services.}\label{fig:unseen}
	\end{subfigure}
	\caption{Distribution of references across services.}
	\label{fig:dist-references}
\end{figure}

For the {\it seen} set in Figure \ref{fig:seen}, we present the distribution of references both in training and test. For the {\it unseen} sets in Figure \ref{fig:unseen}, we present only test reference distribution (since there are no corresponding train references).

\subsection{Performance across Services}
Table \ref{table:results-domains} shows the performance of each model across all seen and partially-unseen test sets.

\begin{table*}[t!]
	\begin{footnotesize}
		
		\begin{center}
			\begin{subtable}[t]{\textwidth}
				\begin{tabular}{p{1.5cm}p{2cm}|p{1.5cm}p{1.5cm}|p{1.5cm}p{1.5cm}|p{1.5cm}p{1.5cm}}\toprule
					&              & \multicolumn{2}{|c|}{\sc Seq2Seq}     & \multicolumn{2}{c|}{\sc CVAE}        & \multicolumn{2}{c}{\sc GPT2}        \\\hline
					\bf Service         & \bf \% Test Refs & \bf BLEU  & \bf SER $\downarrow$ & \bf BLEU  & \bf SER $\downarrow$  & \bf BLEU  & \bf SER $\downarrow$ \\\hline
					events\_1      & 19\%         & 0.6168   & 0.0490  & 0.6126     &  0.0294  & 0.4682   & 0.0588  \\
					rentalcars\_1  & 18\%         &  \bf 0.7486  & 0.1500  & 0.6645   &  0.1125  & 0.6173   & 0.1000  \\
					buses\_1       & 15\%         &  0.3831  & 0.1542  & \it 0.5035    & 0.1000  &  0.4016   & 0.0167  \\
					
					homes\_1       & 9\%          &\it  0.3672   & 0.0660  &  0.5132    & 0.1176  & 0.4836  & 0.0065  \\
					ridesharing\_1 & 9\%          & 0.6334  &  \it 0.2292  & 0.6316    &  \it 0.1667  & 0.6288  &  \bf 0.0000  \\
					hotels\_1      & 8\%          & 0.4414   & 0.0983  & 0.5094    & 0.0700  & \it  0.3405   &  \bf 0.0000  \\
					music\_1       & 8\%          & 0.6807  & 0.1111  & \bf 0.8538 &  \bf 0.0278  &  \bf 0.6961  &  \bf 0.0000  \\
					travel\_1      & 7\%          & 0.4542   & \bf 0.0175  & 0.4334  &   0.1053  & 0.3762  &  \bf 0.0000  \\
					weather\_1     & 7\%          & 0.6302   & 0.1528  & 0.7578 & 0.1111  & 0.5830   &  \it 0.1667  \\
				\end{tabular}
				\caption{\label{table:seen-app} Seen services.}
			\end{subtable}
		\end{center}
		\vfill
		
		\begin{center}
			\begin{subtable}[t]{\textwidth}
				\begin{tabular}{p{1.5cm}p{2cm}|p{1.5cm}p{1.5cm}|p{1.5cm}p{1.5cm}|p{1.5cm}p{1.5cm}}\toprule
					&              & \multicolumn{2}{|c|}{\sc Seq2Seq}     & \multicolumn{2}{c|}{\sc CVAE}        & \multicolumn{2}{c}{\sc GPT2}        \\\hline
					\bf Service         & \bf \% Test Refs & \bf BLEU  & \bf SER  $\downarrow$& \bf BLEU  & \bf SER $\downarrow$ & \bf BLEU  & \bf SER $\downarrow$ \\\hline
					restaurants\_2 & 24\% & \it 0.2466   &  \bf 0.2098  &  \it 0.2126   & 0.3501  &  \it 0.2297   &  \bf 0.0527  \\
					flights\_3     & 18\%         &   0.3193  &  \it 0.4579  & 0.3481   &  \it 0.5000  & 0.3008    & 0.7368  \\
					services\_4    & 18\%         &  \bf 0.5791  & 0.2197  & 0.3288   & 0.4013  &  \bf 0.5760 & 0.0851  \\
					hotels\_4      & 17\%         & 0.3601  & 0.2284  & 0.3381   & 0.2978  & 0.4173  & 0.1552  \\
					banks\_2       & 10\%         & 0.4305   & 0.2546  &  \bf 0.4578   &  \bf 0.2315  & 0.5049  & 0.3519  \\
					media\_2       & 9\%          & 0.3914   & 0.3218  & 0.3815   & 0.3218  & 0.3249  & 0.4483  \\
					movies\_2      & 4\%          & 0.3956  & 0.4028  & 0.3556 & 0.4444  & 0.3800  &  \it 0.8472  \\
				\end{tabular}
				\caption{\label{table:partially-unseen-app} Partially-unseen services.}
			\end{subtable}
		\end{center}
	\end{footnotesize}
	\vspace{-.1in}
	\caption{\label{table:results-domains-app} Automatic evaluation metrics across seen and partially-unseen services (best in bold, worst in italic).}
\end{table*}

\section{Seq2Seq and CVAE Model Diagram}
\begin{figure*}[h!]
	\includegraphics[width=0.9\linewidth]{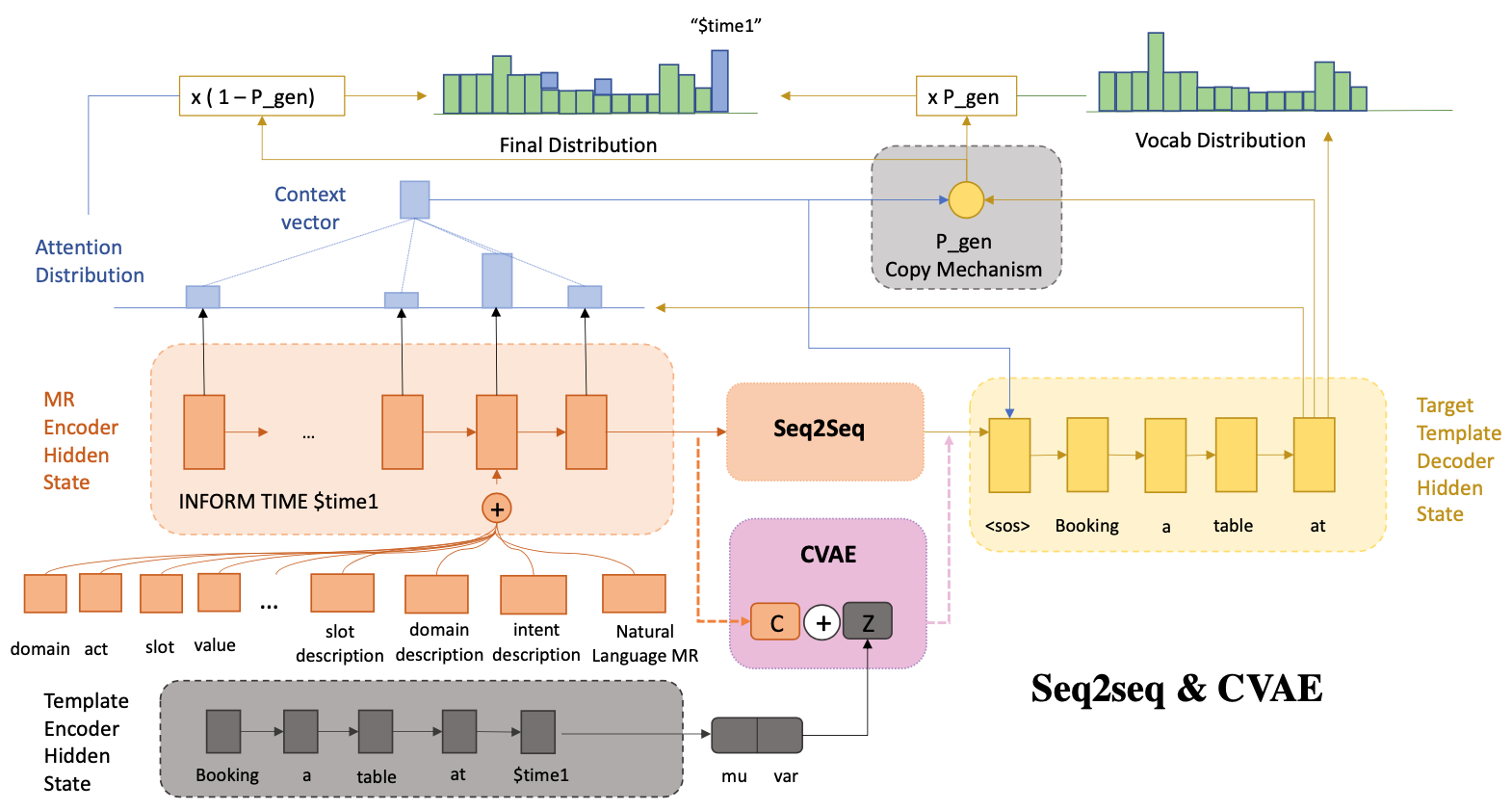}
	\caption{Seq2seq and CVAE model architectures}\label{fig:seq2seq_cvae}
\end{figure*}

Figure \ref{fig:seq2seq_cvae} shows a model architecture diagram for Seq2Seq and CVAE. 

\newpage
\section{Output Examples}
Table \ref{table:more-output-examples} shows more model output examples. Schema 1 shows correct outputs for all models. Schema 2 shows a slot drop in CVAE, and Schema 3 shows incorrect outputs from Seq2Seq/CVAE for the single fully-unseen domain, {\it alarm-1}.

\begin{table*}[t!]
	\begin{small}
		\begin{tabular}{p{1cm}|p{14cm}}\toprule
			\rowcolor{lightgray}	
			\multicolumn{2}{l}{{\bf [Schema 1] } \it \underline{ACTIONS (MR):} CONFIRM(leaving-date=\$leaving-date1), CONFIRM(travelers=\$travelers1)} \\
			\rowcolor{lightgray}	
			\multicolumn{2}{l}{\it \underline{Slot Desc:} leaving-date: "date of bus leaving for journey", travelers: "number of travelers for journey" } \\
			\rowcolor{lightgray}
			\multicolumn{2}{l}{\it \underline{Service:} buses-1 $\:\:\:\:\:\:\:\:\:\:$  \underline{Service Desc:} book bus journeys from the biggest bus network in the country}\\
			\rowcolor{lightgray}
			\multicolumn{2}{l}{\it \underline{Intent:} BuyBusTickets $\:\:\:\:$ \underline{Intent Desc:} buy tickets for a bus journey} \\
			\rowcolor{lightgray}
			\multicolumn{2}{l}{\it \underline{Natural Language MR:} please confirm that the leaving date is \$leaving-date1. please confirm that the travelers is \$travelers1.} \\
			\hline
			\rowcolor{LightCyan}
			\it Ref & \it can you confirm once again that you need tickets for \$travelers1 people for the bus leaving on \$leaving-date1. \\
			Seq2Seq & please confirm the following details : you want to book \$travelers1 tickets on \$leaving-date1 .\\
			CVAE & please confirm : \$travelers1 tickets for the bus leaving on \$leaving-date1 .\\
			GPT2 & okay, it's \$travelers1 tickets leaving \$leaving-date1, is that right?\\ \hline
			
			\addlinespace[2ex] \hline
			\rowcolor{lightgray}
			\multicolumn{2}{l}{{\bf [Schema 2] } \it \underline{ACTIONS (MR):} INFORM(ride-fare= \$ride-fare1), INFORM(approx-ride-duration=\$approx-ride-duration1)} \\
			\rowcolor{lightgray}
			\multicolumn{2}{l}{\it \underline{Slot Desc:} ride-fare"total fare for taxi ride",
				approx-ride-duration: "approximate duration of ride to the destination"} \\
			\rowcolor{lightgray}
			\multicolumn{2}{l}{\it \underline{Service:} ridesharing-1$\:\:\:\:\:\:\:\:\:\:$  \underline{Service Desc:} on-demand taxi calling service}\\
			\rowcolor{lightgray}
			\multicolumn{2}{l}{\it \underline{Intent:} GetRide $\:\:\:\:\:\:\:\:\:\:$ \underline{Intent Desc:} call a taxi to head to a given destination} \\
			\rowcolor{lightgray}
			\multicolumn{2}{l}{\it \underline{Natural Language MR:} the ride fare is [\$ride-fare1]. the approximate ride duration is [\$approx-ride-duration1].} \\
			\hline
			\rowcolor{LightCyan}
			\it Ref & \it around \$approx-ride-duration1 minutes for \$ride-fare1. \\
			Seq2Seq & around \$approx-ride-duration1 minutes for \$ride-fare1.\\
			CVAE & the cab is on the way . it will cost \$ride-fare1.\\
			GPT2 &the ride will be \$ride-fare1 and is expected to last \$approx-ride-duration1 minutes. \\ \hline
			
			\addlinespace[2ex] \hline
			\rowcolor{lightgray}
			\multicolumn{2}{l}{{\bf [Schema 3] } \it \underline{ACTIONS (MR):} REQUEST(new-alarm-time=null)} \\
			\rowcolor{lightgray}
			\multicolumn{2}{l}{\it \underline{Slot Desc:} new-alarm-time: "time to set for the new alarm" } \\
			\rowcolor{lightgray}
			\multicolumn{2}{l}{\it \underline{Service:} alarm-1 $\:\:\:\:\:\:\:\:\:\:$  \underline{Service Desc:} Manage alarms by getting and setting them easily}\\
			\rowcolor{lightgray}
			\multicolumn{2}{l}{\it \underline{Intent:} AddAlarm $\:\:\:\:$ \underline{Intent Desc:} Set a new alarm} \\\rowcolor{lightgray}
			\multicolumn{2}{l}{\it \underline{Natural Language MR:} what [new alarm time] do you want?} \\
			\hline
			\rowcolor{LightCyan}
			\it Ref & \it at what time, you want the alarm to be set? \\
			Seq2Seq &what time do you want to go ? \\
			CVAE & what time would you like to go ?\\
			GPT2 & what time do you want to set the alarm?\\ \hline
			
		\end{tabular}
	\end{small}
	\vspace{-.1in}
	\centering \caption{\label{table:more-output-examples} Example model outputs.} 
\end{table*} 

\end{document}